\documentclass[runningheads]{llncs}
\usepackage{graphicx}
\usepackage{amsmath,amssymb} 
\usepackage{color}
\usepackage[width=122mm,left=12mm,paperwidth=146mm,height=193mm,top=12mm,paperheight=217mm]{geometry}

\usepackage{ifthen}
\usepackage{adjustbox}
\usepackage{epsfig}
\usepackage{array}
\usepackage{graphicx}
\usepackage{sidecap}
\usepackage[pagebackref=true,breaklinks=true,letterpaper=true,colorlinks,bookmarks=false]{hyperref}
\definecolor{gray}{rgb}{0.5,0.5,0.5} 
\definecolor{green}{rgb}{0, 0.4, 0} 
\definecolor{orange}{rgb}{1, 0.5, 0} 	
\definecolor{mahogany}{rgb}{0.75, 0.25, 0.0}
\definecolor{purple}{rgb}{0.6, 0, 0.6}
\definecolor{purple}{rgb}{0.6, 0, 0.6}
\definecolor{darkgreen}{rgb}{0, 0.4, 0} 
\definecolor{frenchblue}{rgb}{0.0, 0.45, 0.73}

\newboolean{revising}
\setboolean{revising}{false}
\ifthenelse{\boolean{revising}}
{
	\newcommand{\ignore}[1]{}
	\newcommand{\yu}[1]{\textcolor{black}{#1}}
	
	\newcommand{\lu}[1]{\textcolor{black}{#1}}

	\newcommand{\tingfan}[1]{\textcolor{black}{#1}}
	
	\newcommand{\sunmin}[1]{\textcolor{black}{#1}}
	
    \newcommand{\highlight}[1]{\textcolor{red}{#1}}
} {
	\newcommand{\ignore}[1]{}
	\newcommand{\yu}[1]{#1}
	
	\newcommand{\lu}[1]{#1}

	\newcommand{\tingfan}[1]{#1}
	
	\newcommand{\sunmin}[1]{#1}
	
	\newcommand{\highlight}[1]{#1}
}

\begin{document}
\pagestyle{headings}
\mainmatter

\title{Leveraging Motion Priors in Videos for Improving Human Segmentation} 

\titlerunning{Leveraging Motion Priors in Videos}

\authorrunning{Y.-T. Chen, W.-Y. Chnag, H.-L. Lu, T. Wu, and M. Sun }

\author{Yu-Ting Chen\inst{1}, Wen-Yen Chang\inst{1}, Hai-Lun Lu\inst{1}, Tingfan Wu\inst{2}, Min Sun\inst{1}}


\institute{National Tsing Hua University
\email{\{yuting2401,s0936100879,oscar.lu1007\}@gmail.com, sunmin@ee.nthu.edu.tw}\\
\and
Umbo Computer Vision\\
\email{tingfan.wu@umbocv.com}}

\maketitle

\begin{abstract}
Despite many advances in deep-learning based semantic segmentation, performance drop due to distribution mismatch is often encountered in the real world. Recently, a few domain adaptation \yu{and active learning approaches} have been proposed to mitigate the performance drop. However, very little attention has been made toward leveraging information in videos which are naturally captured in most camera systems.
\yu{In this work, we propose to leverage ``motion prior'' in videos for improving human segmentation \sunmin{in a weakly-supervised active learning setting.}}
By extracting motion information using optical flow in videos, we can extract candidate foreground motion segments (referred to as motion prior) potentially corresponding to human segments. 
We propose to learn a  memory-network-based policy model to select \textit{strong} candidate segments (referred to as \textit{strong} motion prior) through reinforcement learning. The selected segments have high precision and are \sunmin{directly used to finetune the model}.
In a newly collected surveillance camera dataset and a publicly available UrbanStreet dataset, our proposed method improves the performance of human segmentation across multiple scenes and modalities (i.e., RGB to Infrared (IR)). Last but not least, our method is empirically complementary to existing domain adaptation approaches such that additional performance gain is achieved by combining our weakly-supervised active learning approach with domain adaptation approaches.

\end{abstract}

\section{Introduction} \label{sec:intro}


Intelligent camera systems with the capability to recognize objects often encounter issues caused by data distribution mismatch in the real world. For instance, surveillance cameras encounter various weather conditions, 
view angles, lighting conditions, and sensor modalities (e.g., RGB, infrared or even thermal). A standard solution is to collect more labeled images from various distributions to train a more robust model. However, collecting high-quality
labels is very expensive and time-consuming, especially for segmentation and detection tasks. 
These considerations raise two critical questions: (1) ``how to select data points for training such that the accuracy improved as much as possible?" and (2) ``how to obtain the label of the selected data points with cost as low as possible?''

Active learning is one of the common paradigms to address the ``how to select'' question since it is defined as learning to select data points to label, from a pool of unlabeled data points, in order to maximize the accuracy. There exist many heuristics~\cite{Active2010} which have been proven to be effective when applied to classical machine learning models. However, Sener and Savarese~\cite{sener2017geometric} have shown that these heuristics are less effective when applied to CNN. To overcome the limitation, Sener and Savarese~\cite{sener2017geometric} propose a new active learning method specifically designed for Convolutional Neural Networks (CNNs). Despite recent advances, Most active learning methods require human to label the selected data points. For segmentation and detection tasks, the cost of labeling a small set of selected data points can still be relatively expensive and time-consuming.

On the other hand, instead of collecting independent images, it is generally easy to collect a sequence of images (i.e., a video) from always-on camera systems.   Sequences of images have two main properties: (1) images close in time are similar/redundant, and (2) difference in two consecutive images reveals motion information potentially corresponding to moving objects.
Very little attention, however, has been made toward exploiting these properties in a video to automatically provide supervision to boost recognition performance and mitigate the performance drop caused by distribution mismatch. 
This is related to the ``how to obtain labels'' question. 
If we can obtain labels automatically from videos, it will be immensely beneficial for intelligent camera systems.
In fact, researchers have proposed to extract motion information from a sequence of images. For instance, given two consecutive frames, dense optical flow can be extracted for each pixel. Given a longer sequence of frames, sparse long-term trajectories of pixels can be extracted. In the rest of the paper, we refer to these motion information in a video as ``motion prior''.




\begin{figure}[t!]
\begin{minipage}{0.6\textwidth} 
\includegraphics[width=0.92\textwidth]{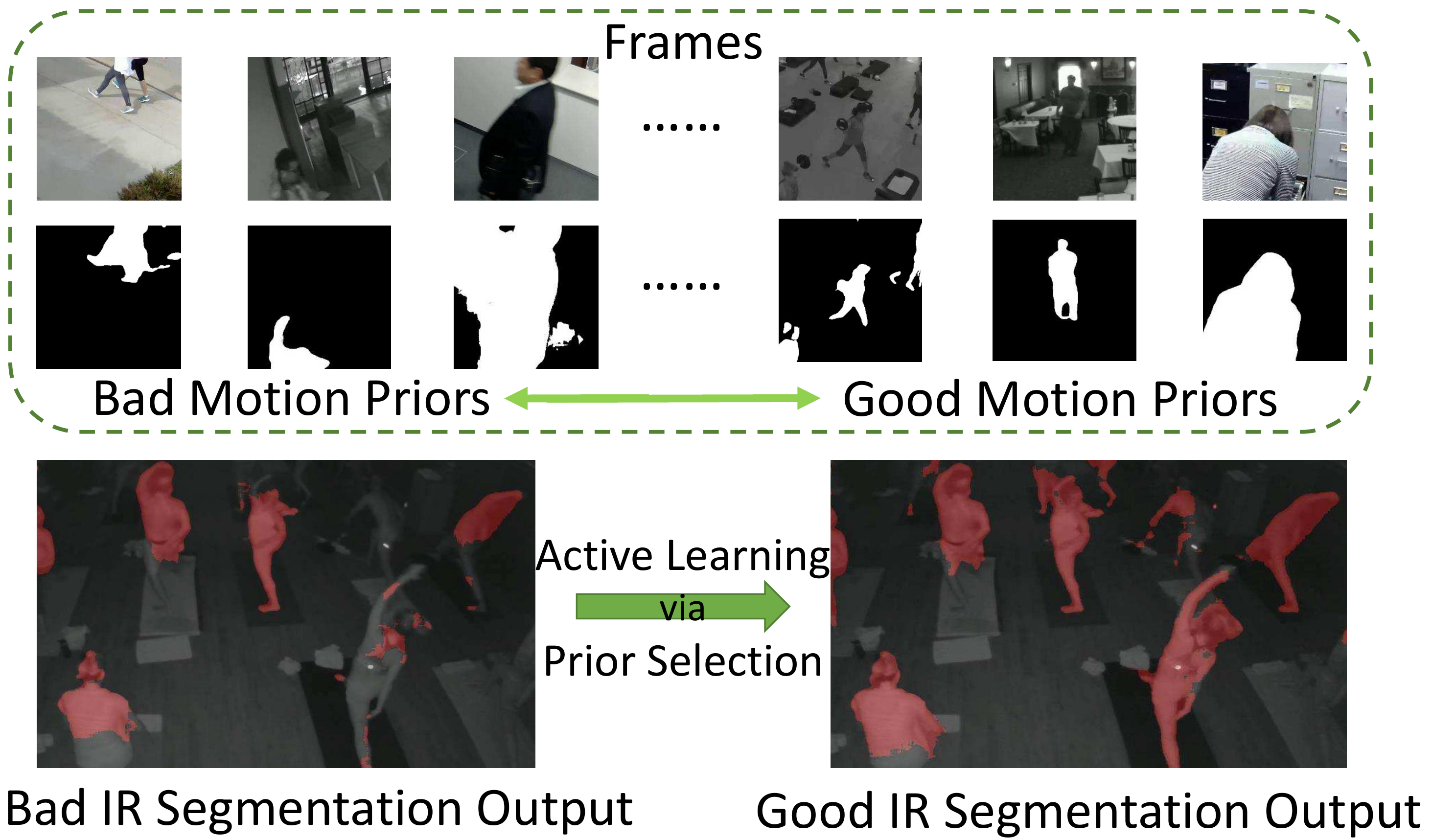}\label{fig:mask}
\end{minipage}%
\begin{minipage}{0.4\textwidth}
\caption{\sunmin{(top): RGB patches and their corresponding patch-based motion priors extracted from videos. The priors can be classified into ``good" and ``bad" ones. (bottom): Our proposed active learning strategy can select good motion priors to improve performance in a cross-modality (RGB to IR) segmentation scenario. 
}}
\end{minipage}
\end{figure}



In this work, we propose to leverage 
motion prior in videos for improving human segmentation accuracy. We first compute dense optical flow between two consecutive frames. Then, we treat pixels with flow higher than a threshold as candidates of foreground motion segments, which are referred to as ``motion prior''. Due to the nature of imperfect optical flow, a majority of the segments are quite noisy (see examples in Fig.~\ref{fig:mask}).
Considering that only some candidates are good and many candidates are noisy, we propose to learn a memory-network-based policy model to select good candidate segments through reinforcement learning. The selected good segments are then used as additional ground truth to finetune the human segmenter. 
In this way, we can achieve active learning without additional human annotation.

Our policy is trained on a hold-out dataset with unlabeled videos and a set of labeled images. The training of the policy is formulated as a reinforcement learning problem where the reward is the accuracy of the labeled images and the action is whether to select each motion segment. Once the policy is trained, we can apply the policy to select motion segments in challenging cross-modality (RGB to IR) or cross-scene settings. We refer our setting as weakly-supervised active learning since the policy needs to be trained on an additional hold-out dataset. 

In a newly collected surveillance camera dataset and a publicly available UrbanStreet dataset, our proposed method improves the performance of human segmentation across multiple scenes and modalities (i.e., RGB to Infrared (IR)). Last but not least, our method is empirically complementary to existing domain adaptation approaches such that additional performance gain is achieved by combining our weakly-supervised active learning approach with domain adaptation approaches.




In the following sections, we first describe the related works in Sec.~\ref{sec:related}. Then, we introduce our new surveillance cameras dataset in Sec.~\ref{sec:data}. Our main technical contribution---policy-based weakly-supervised active learning for strong motion prior selection---is introduced in Sec.~\ref{sec:method}. Finally, we report our experimental results in Sec.~\ref{sec:results}.

\section{Related Works} \label{sec:related}
We discuss the related work in the fields of motion segmentation, human segmentation, active learning and domain adaptation.

\subsection{Motion Segmentation} 
Motion segmentation aims to decompose a video into foreground objects and background using motion information. Feature-based motion segmentation methods assume that segmentation of different motions is equivalent to segment the extracted feature trajectories into different clusters. These methods can be classified into two types: affinity-based methods~\cite{dragon2012multi,ochs2014segmentation} and subspace-based  method~\cite{elhamifar2009sparse,yang2017motion}. 
Some of the works utilize properties of trajectory data. For example, Yan and Pollefeys~\cite{yan2006general} use geometric constraint and locality to solve the problem.
Recently,~\cite{tsai2016video,cheng2017segflow} propose to jointly tackle the motion segmentation and optical flow tasks. 
\highlight{Nirkin et al.~\cite{nirkin2018face} use motion as a prior and propose a man in the loop for producing segmentation labels.}
In our work, we simply obtain candidate moving object segments via high-quality optical flow.  
\sunmin{Most importantly, none of the work aforementioned leverage motion segmentation for weakly-supervised active learning.}

\subsection{Human Segmentation}
Human segmentation has a wide range of applications. For instance, \yu{human segmentation in a high-density scene (crowded or occluded) acquired from a stationary camera has been discussed in early works~\cite{zhao2002stochastic,zhao2003bayesian}.} Spina et al.~\cite{spina2013video} demonstrate applications in pose estimation and behavior study. On the other hand, in many applications, real-time performance is critical. Song et al.~\cite{song20151000fps} achieve 1000 fps \lu{using a CNN-based architecture which outperforms traditional methods in both speed and accuracy.}
\highlight{Some works use motion information for helping human segmentation, for instance, Guo et al.~\cite{guo2015video} base on local color distribution and shape priors through optical flow, and Lu et al.~\cite{lu2015human} describe a hierarchical MRF model to bridge low-level video fragments with high-level human motion and appearance.}

In recent years, thermal \tingfan{and} infrared systems have gained popularity for night vision. Hence, human segmentation on infrared images has become an important topic. For example, Tan et al.~\cite{tan2013background} propose a background subtraction based method for human segmentation on thermal infrared images. He et al.~\cite{he2017human} further utilize predicted human segments on infrared images to guide robots search. To demonstrate severe domain shift, we evaluate our method mainly on cross-modality (RGB to IR) domain adaptation for human segmentation.

\subsection{Active learning}
An active learning algorithm can explore informative instances, querying desired output form users or other sources. 
Uncertainty-based approaches are widely used. These works consider uncertainty as the selection strategies. They find hard examples by dropout MC sampling~\cite{yarin2017deep}, using heuristics like highest entropy~\cite{joshi2009multi}, or geometric distance to decision boundaries~\cite{brinker2003incorporating,melanie2018adversarial}. 
Other approaches consider the diversity of selected samples, using k-means algorithms~\cite{sener2017geometric,li2012incorporating} or sparse representation for subset selection~\cite{ehsan2013convex}. Still other important concepts also help the performance, such as selecting instances which will maximize the variance of output~\cite{yazhou2017variance,christoph2016large}, or introducing the relationships between data points in structured data~\cite{ankit2011relational,sujoy2017non_uniform}.

Recently, some works model the active learning process as a sequence of querying actions, using deep reinforcement learning. Fang et al.~\cite{fang2017learning} demonstrates on cross-lingual setting and Bachman et al.~\cite{bachman2017learning} models the learning algorithm via meta-learning. Our approach is similar to these methods using learnable strategy rather than predefined heuristic.  Above methods show their goal to reduce human label cost. \sunmin{However, we use active learning for unsupervised finetuning since our method selects automatically computed motion priors, requiring ZERO human label cost once the policy has learned.}

\subsection{Domain Adaptation}
Domain adaptation leverages information from one or more source domains to improve the performance on target domain. Recent methods focus on learning deep  representations to be robust to domain shift~\cite{tzeng2015simultaneous}.
Several other works propose to align source and target domains in feature space based on Maximum Mean Discrepancy (MMD)~\cite{long2015learning} or Central Moment Discrepancy (CMD)~\cite{zellinger2017central}.

On the other hand, adversarial training\yu{~\cite{goodfellow2014generative}} has been applied for domain adaptation as well\yu{~\cite{liu2016coupled,ganin2015unsupervised,tzeng2017adversarial}}. Liu et al.~\cite{liu2016coupled} propose Coupled GAN which generates a joint distribution of two domains for classification. 
Ganin et al.~\cite{ganin2015unsupervised} applies adversarial training for achieving maximal confusion between the two domains.
Other works such as Domain Separation Networks (DSN)~\cite{bousmalis2016domain} split the feature into shared representations and private ones, in order to improve the ability to extract domain-invariant features. Most of the works mentioned above focus on classification. Hoffman et al.~\cite{hoffman2016fcns}, Chen et al.~\cite{chen2017no} and more recent works~\cite{zhang2017curriculum,sankaranarayanan2017unsupervised} extend to segmentation which is closer to our human segmentation task. 
\sunmin{In this work, we show that our proposed weakly-supervised active learning approach is complementary to state-of-the-art domain adaptation approaches.}








\def \Ltotal {\mathcal{L}_{total}}
\def \Ltask {\mathcal{L}_{task}}
\def \Lga {\mathcal{L}_{global}}
\def \Lca {\mathcal{L}_{class}}
\def \Lsel {\mathcal{L}_{T}}
\def \lmga {\lambda_{G}}
\def \lmca {\lambda_{C}}
\def \lmsel {\lambda_{T}}
\def \Lp {\mathcal{L}_{Policy}}
\def \Ld {\mathcal{L}_{D}}
\def \Ldinv {\mathcal{L}_{Dinv}}
\def \Lda {\mathcal{L}_{DA}}
\def \lmda {\lambda_{D}}
\def \Mf {\varphi_F}
\def \My {\varphi_Y}
\def \Mga {\varphi_{D}}
\def \Mca {\varphi_{D}^c}

\def \thf {\theta_F}
\def \thy {\theta_Y}
\def \thga {\theta_{D}}
\def \thca {\theta_{D}^{c}}

\def \thp {\theta_P}


\def \S  {\mathcal{S}} 
\def \St  {\mathcal{S}_\mathcal{T}} 
\def \Se  {\mathcal{S}_\mathcal{E}} 

\def \T  {\mathcal{T}} 
\def \Tt  {\mathcal{T}_\mathcal{T}} 
\def \Te  {\mathcal{T}_\mathcal{E}} 

\def \Is {I^\mathcal{S}} 
\def \It {I^\mathcal{T}} 
\def \Vs {V^\mathcal{S}} 
\def \Vt {V^\mathcal{T}} 

\def \Ist {I^\mathcal{S}_\mathcal{T}} 
\def \Ise {I^\mathcal{S}_{E}} 
\def \Itt {I^\mathcal{T}_\mathcal{T}} 
\def \Ite {I^\mathcal{T}_{E}} 

\def \Id {I} 
\def \Xs {X_\mathcal{S}} 
\def \Xt {X_\mathcal{T}} 


\def \fx {\Mf(x,\thf)} 
\def \fs {\Mf(\Is,\thf)}
\def \ft {\Mf(\It,\thf)}

\def \Ys  {Y_S}   
\def \Yt  {Y_T}   
\def \mi  {m_i}   
\def \phii {\phi^c_i}
\def \Phin {\Phi^c_n}
\def \tPhin {\tilde{\Phi}^c_n}
\def \phiih {\phi^{\hat{c}}_i}
\def \tphii {\tilde{\phi}^c_i}

\def \pn {p_n} 
\def \pcn {p^c_n} 

\def \Rn {\mathcal{R}_n} 

\def \C {\mathcal{C}} 

\def \E  {\mathcal{E}} 
\def \phip {\phi} 
\def \phiseg {\phi_{S}} 
\def \phisego {\phi_{S}^0} 

\def \gcn {g^c_n} 
\def \gcnhat {\hat{g}^{c}_n} 
\def \moin {m(\Ist(R_n))} 
\def \imin {\Ist(R_n)} 
\def \moink {m(\Ist(R_{n,k}))} 
\def \imink {\Ist(R_{n,k})} 

\section{Surveillance Datasets} \label{sec:data}
 In order to create challenging scenarios in videos, we have collected a new surveillance camera dataset consisting of large distribution mismatch due to cross-domains scenarios: cross-modalities (i.e., RGB to InfraRed (IR)) and across-scenes. \tingfan{It is surprisingly difficult to find existing segmentation annotated cross-domains video dataset.   Due to the high cost of labeling, most public annotated video dataset are usually very small, not to mention about crossing multiple domains. In our dataset; we highlight cross-modalities for its high appearance mismatch and practical value.  
 For surveillance application,  good human segmentations across multiple sensor modality and scenes is essential. This dataset directly validates the proposed method in real-world surveillance scenarios. 
}

We collect four datasets: Gym-RGB, Gym-IR, Store-RGB \tingfan{, and} Multi-Scene-IR. There are two different sensor modes on typical surveillance cameras, color and infrared, which we denote as ``RGB'' and ``IR'', respectively. \tingfan{To simulate real-world usage, we let the camera ambient light sensor to automatically switch between the two modes. Typically, when there is sufficient lighting, the cameras operate in RGB mode; on the other hand, when it gets dark, the IR mode is activated to improve sensitivity.}
All datasets are videos collected by stationary cameras,  we label a subset of frame sparsely sampled from each video. 

\subsection{Cross-domains Settings}
We divide our data into source $\mathcal{S}$ and target $\mathcal{T}$ domains. In this dataset, we treat all RGB data as source domain and all IR data as target domain in order to test challenging cross-modalities settings. In both domains, we further define training $T$ and evaluation $E$ sets. All evaluation set contains labeled images.
In the source domain, training $T$ consists of a few labeled images $\mathcal{I}^\mathcal{S}_T$ and unlabeled video frames $\mathcal{V}^\mathcal{S}_T$. The labeled training images $\mathcal{I}^\mathcal{S}_T$ are used to pre-train our segmenter. The unlabeled video frames $\mathcal{V}^\mathcal{S}_T$ are used to extract motion prior information (Sec.~\ref{priors}). Both the unlabeled video frames $\mathcal{V}^\mathcal{S}_T$ and the evaluation set $\mathcal{I}^\mathcal{S}_E$ in the source domain are used to train our motion prior selector using reinforcement learning (Sec.~\ref{SEL}). 
In the target domain, training $T$ consists of only unlabeled video frames $\mathcal{V}^\mathcal{T}_T$ which are used to extract motion prior information.
Finally, we report the cross-domains performance on the evaluation set $\mathcal{I}^\mathcal{T}_E$ in the target domain.
\sunmin{The statistics about a number of videos and labeled images in each set of the source and target domain are shown in Table.~\ref{table:source_data} and~\ref{table:target_data}, respectively.}
\newcolumntype{L}{@{\hskip\tabcolsep\vrule width 1.3pt\hskip\tabcolsep}}

\begin{table}[h]
\begin{minipage}{0.55\textwidth} 
\begin{tabular}{|c|c|cLc|c|c|}
\hline
 \multicolumn{3}{|cL}{{Gym-RGB}} & \multicolumn{3}{c|}{{Store-RGB}} \\
\hline
 \multicolumn{2}{|c|}{{Train}} & {Test} & \multicolumn{2}{c|}{{Train}} & {Test} \\
 \hline
 {Images} & {Videos} & {Images} & {Images} & {Videos} & {Images} \\
 \hline
 749 & 406 & 237 & 985 & 985 & 255\\ 
\hline
\end{tabular}
\end{minipage}
\begin{minipage}{0.45\textwidth} 
\caption{{Source domain datasets. ``Images" refers to the number of images that are labeled. ``Videos" refers to the number of videos that contain unlabeled frames.}}
\label{table:source_data}
\end{minipage}
\end{table}

\newcolumntype{K}[1]{>{\centering\arraybackslash}p{#1}}
\begin{table}[h]
\begin{minipage}{0.55\textwidth} 
\begin{tabular}{|K{1.52cm}|K{1.52cm}LK{1.52cm}|K{1.52cm}|}
\hline
 \multicolumn{2}{|cL}{{Gym-IR}} & \multicolumn{2}{c|}{{Multi-Scene-IR}} \\
\hline
 {Train} & {Test} & {Train} & {Test} \\
 \hline
{Videos} &{Images} & {Videos} & {Images} \\
 \hline
 929 & 492 & 253 & 89\\ 
\hline
\end{tabular}
\end{minipage}
\begin{minipage}{0.45\textwidth}
\caption{{Target domain datasets. ``Images" refers to the number of labeled images. ``Videos" refers to the number of videos consist of unlabeled frames. Note that there are no labeled training images in the target domain.}}
\label{table:target_data}
\end{minipage}
\end{table}
\subsection{Data Collection Details}
For the Store-RGB dataset, we have only color (RGB) images since there is sufficient fluorescent lighting in the stores all day. On the other hand, we collect infrared data (Multi-Scene-IR) from multiple scenes, such as house, office, walkway, park, playground, etc. \tingfan{For Gym scene, the data comes in both RGB and IR modalities due to natural day-and-night lighting transitions.} For all videos, there are about 6 to 15 frames in one video with 1080$\times$1920 resolution.

\section{Our Method} \label{sec:method}

We describe how to obtain motion prior from optical flow  (Sec.~\ref{priors}) and select a set of \textit{strong} motion prior. 
Before that, we first define some common notations below. 

\noindent\textbf{Notation.} 
We use $i$, $n$, and $k$ to index pixel, patch and the order of input data, respectively. \textbf{m} indicates motion prior, and $m_i$ denotes the motion prior of the $i^{\text{th}}$ pixel.   


\subsection{Motion Priors from Video Frames}\label{priors}
Our goal is to obtain a set of motion prior $\textbf{m}$ (i.e., candidate foreground mask) from video frames.
Although many sophisticated motion segmentation methods can be used, we simply apply a state-of-the-art optical flow method~\cite{ilg2016flownet}. Then, we obtain $\textbf{m}$ as the binarized flow map such that $m_i=1$ if its flow magnitude is larger than a threshold $\tau$. Since surveillance cameras in our dataset are typically stationary, we may assume that most background and foreground pixels corresponding to small and large flow magnitude, respectively. For non-stationary cameras, other motion segmentation methods (e.g.,~\cite{motionEst}) can be used to handle camera motion.

These automatically obtained motion priors inevitably will be noisy and contain outliers. Hence, we propose a memory-network-based policy model to select more accurate ones instead of directly finetuning the segmenter with all noisy labels. The usage of motion priors is illustrated in Fig.~\ref{finetune-tgt}.

\subsection{Motion Priors Selection}\label{SEL}

\begin{figure}
\begin{center}
\includegraphics[width=0.86\columnwidth]{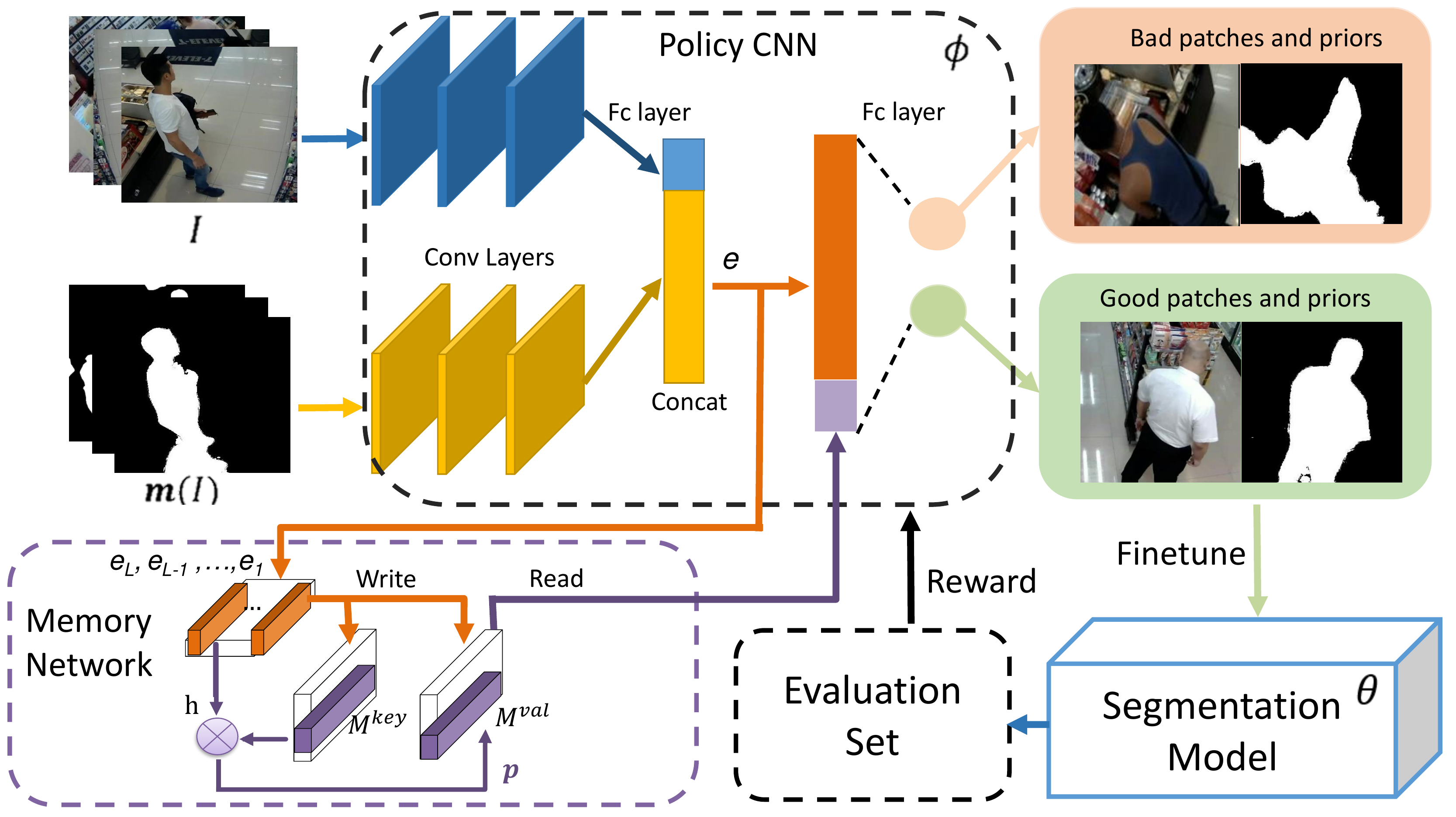}
\end{center}
\caption{Training Procedure of Policy Model via reinforcement learning. The policy model $\phi$ (consist of policy CNN and memory network) takes both the image $I$ and the motion prior $\textbf{m}(I)$ as inputs and predicts an action, selecting $\textbf{m}(I)$ as a good prior or not. The selected priors are further used to improve segmenter $\theta$, and then the improvement shown on a hold-out evaluation set will become a reward to update the policy model $\phi$.}
\label{fig:policy}
\end{figure}

We train a policy model $\pi$ which learns to select a set of \textit{strong} motion priors. Further, these \textit{strong} motion priors are treated as ground truth to directly fine-tune our model using cross-entropy loss. Instead of manually labeling \textit{strong} motion priors and training the policy in a supervised fashion, we train the policy using reinforcement learning, which rewards from directly improving the human segmentation accuracy on a hold-out evaluation set in source domain. The training procedure of our policy model is illustrated in Fig.~\ref{fig:policy}.

\noindent\textbf{Policy model.}
We define the policy $\pi$ as the following probability function:
\begin{eqnarray}
\pi(a|I,\textbf{m}(I);\phi)~,
\end{eqnarray}
where $I$ is an image, $\textbf{m}(I)$ is its corresponding motion prior, $a\in \{0,1\}$ is the binary action to select ($a=1$) or not ($a=0$), and $\phi$ is the model parameters.

\subsubsection{Network Architecture.}

\yu{Inspired by the ideal using Memory Network~\cite{Weston2015memory} in Deep Q-Network (DQN) proposed by Oh at el.~\cite{oh2016control}, we use an memory-network-based policy model which consists of three components: (1) a feature encoder for extracting features from images and motion priors, (2) a memory retaining a recent history of observations, and (3) an action decision layers taking both content features and retrieved memory state to decide the action.}

\noindent\textbf{Feature encoder.}
We propose a two-stream CNN to firstly encode image appearance $I$ and motion prior $\textbf{m}(I)$ separately. To fuse them, we concatenate the embedded features from two streams. Then, we apply a linear transformation on the concatenated feature to mix the features. 
\highlight{Not that it is essential to make our policy network robust to domain shift since it is trained only in source domain but applied in the target domain. We found motion priors are more invariant (relative to RGB images) across domains. Hence, we propose late-fusion and increase the number of features for motion priors.}

\noindent\textbf{Memory network.}
There are two operations, ``write'' and ``read'', in memory network, which is similar to the architecture proposed in~\cite{oh2016control}. 

\begin{itemize}
\item Write.  

The encoded features of last $L$ observations are stored into the memory by linear transformation. Two types of memories are represented as \textit{key} and \textit{value}, which are defined as follows,
\begin{eqnarray}\label{write}
{M}^{key}_{k} = {W}^{key} {E}_{k} \\
{M}^{val}_{k} = {W}^{val} {E}_{k} ,
\end{eqnarray}
where ${M}^{key}_{k}$, ${M}^{val}_{k}$ $\in \mathbb{R}^{d\times L}$ are stored memories with embedding dimension $d$, and $k$ is the index of input data order. ${W}^{key}$ and ${W}^{val}$ are parameters of writing module. ${E}_{k} = \{e_{k-i}\}_{i=1, 2, ..., L}$ $\in \mathbb{R}^{e\times L}$ is concatenation of last $L$ features of observations which are 
selected as good priors.  
\item Read.

Based on soft attention mechanism, the reading output will be the inner product between the content embedding $h$ and key memories ${M}^{key}_{k}$.

\begin{equation}\label{read-soft}
p_{k,\ell} = \frac{\exp({h^\intercal_k} M^{key}_k[\ell])}{\sum^L_{j=1}\exp({h^\intercal_k} M^{key}_k[j])}, ~
\end{equation}

where $h_k = W^h e_k$, and $W^h$ are model parameters for content embedding. $p_{k,\ell}$ is the soft attention for $\ell^{th}$ memory block. Take the attention weights on \textit{value} memories ${M}^{val}_{k}$ as the retrieved output, which can be represented as below,
\begin{equation}\label{read-out}
o_k = M^{val}_k p_k,
\end{equation}

where $o_k$ $\in \mathbb{R}^{d}$ is retrieved memory output. 

\end{itemize}
The memory network is expected to handle the problem of data redundancy, or the policy may tend to select very similar candidates. We concatenate the memory output $o_k$ with current content feature $e_k$ as last features for taking action (select or not).
\subsubsection{Reward.}
We use the improved segmentation accuracy on a hold-out set in the source domain as the reward $r$ as follows,
\begin{eqnarray}\label{reward}
r= \text{IoU}(\mathcal{I}^{\mathcal{S}}_{E};\theta)-\text{IoU}(\mathcal{I}^{\mathcal{S}}_{E};\theta^0)~,
\end{eqnarray}
where $\text{IoU}$ is the Intersection over Union (IoU) metric which is standard for semantic segmentation, $\theta^0$ is the initial parameters of the human segmentor, $\theta$ is the current parameters of the human segmentor, and $\mathcal{I}^{\mathcal{S}}_{E}$ is the set of images in the hold-out set in the source domain. 

After few earlier episodes, $\text{IoU}(\mathcal{I}^{\mathcal{S}}_{E};\theta^0)$ is replaced with other estimated baseline value such as averaged reward in near episodes, in order to maintain learning efficiency.

\subsubsection{Policy Gradient.}
According to above reward function, we compute the policy gradient to update the model parameters $\phi$, represented as below,
\begin{eqnarray}\label{surr}
\nabla_\phi \frac{1}{K}\sum\limits_{k=1}^K r \cdot \log \pi(a_k\mid I_k,\textbf{m}(I_k); \phip)~; I_k\in \mathcal{V}^{\mathcal{S}}_{T}~,
\end{eqnarray}\normalsize
where $k$ is the image index, $K = \mid\mathcal{V}^{\mathcal{S}}_{T}\mid$, and $\mathcal{V}^{\mathcal{S}}_{T}$ is the set of unlabelled training video frames in source domain.

\subsubsection{Training Procedure.}

We conduct the following steps iteratively until the reward and policy loss converge.
\begin{itemize}
\item Given $\phi$, we use the policy network to select a set of image (i.e., $\mathcal{K}=\{k;a_k=1\}$) with  motion priors.
\item Given $\mathcal{K}$, we use $(I_k,\textbf{m}(I_k))_{k\in \mathcal{K}}$ as additional pairs of image and ground truth segmentation to finetune the human segmentation parameters $\theta$. 
\item Given the new $\theta$, we compute the reward $r$ in Eq.~\ref{reward}.
\item Given $r$, we compute policy gradient in Eq.~\ref{surr} and update the policy parameters $\phi$ using Gradient Decent (GD).
\item A budget of used data for training the segmenter $\theta$ is defined as $b$, i.e. an episode early stops at step $s$ as $\sum\limits_{k=1}^s{a_k} = b$. Last, we reset the parameters of the segmentor $\theta = \theta^0$ when an episode finishes.
\end{itemize}

We further extend the procedure above from image-based to patch-based selection. We propose to select motion priors at patch-level since there are very few motion priors which are accurate throughout the entire image. In contrast, there are many patch-based motion priors which are almost completely accurate throughout the entire patch.
Next, we define the patch-based selection process.

\subsubsection{Patch-based Selection.}
Define the $n^{\text{th}}$ patch in an image corresponding to a set of pixels $R_n$, we can write patch-based motion prior as,
\begin{eqnarray}
\textbf{m}_n = \{m_i;i\in R_n\}~.
\end{eqnarray}
The image-based policy gradient in Eq.~\ref{surr} is modified to,
\begin{eqnarray}\label{Psurr}
\nabla_\phi \frac{1}{KN}\sum\limits_{k=1}^K \sum\limits_{n=1}^N r \cdot \log \pi(a_{k,n}\mid I_{k,n},\textbf{m}(I_k)_n; \phip)~,
\end{eqnarray}\normalsize
where $I_{k,n}$ denotes the appearance of the $n^{\text{th}}$ patch on the $k^{\text{th}}$ image, $N$ is the number of patches in an image.
In order to focus on foreground patches and reduce search space, we also automatically filter out patches with all background motion prior (i.e., $m_i=0$ for all $i\in R(n)$).

\subsubsection{Inference on Target Domain.} 
\begin{figure*}[!t]
\begin{center}
\includegraphics[width=0.86\columnwidth]{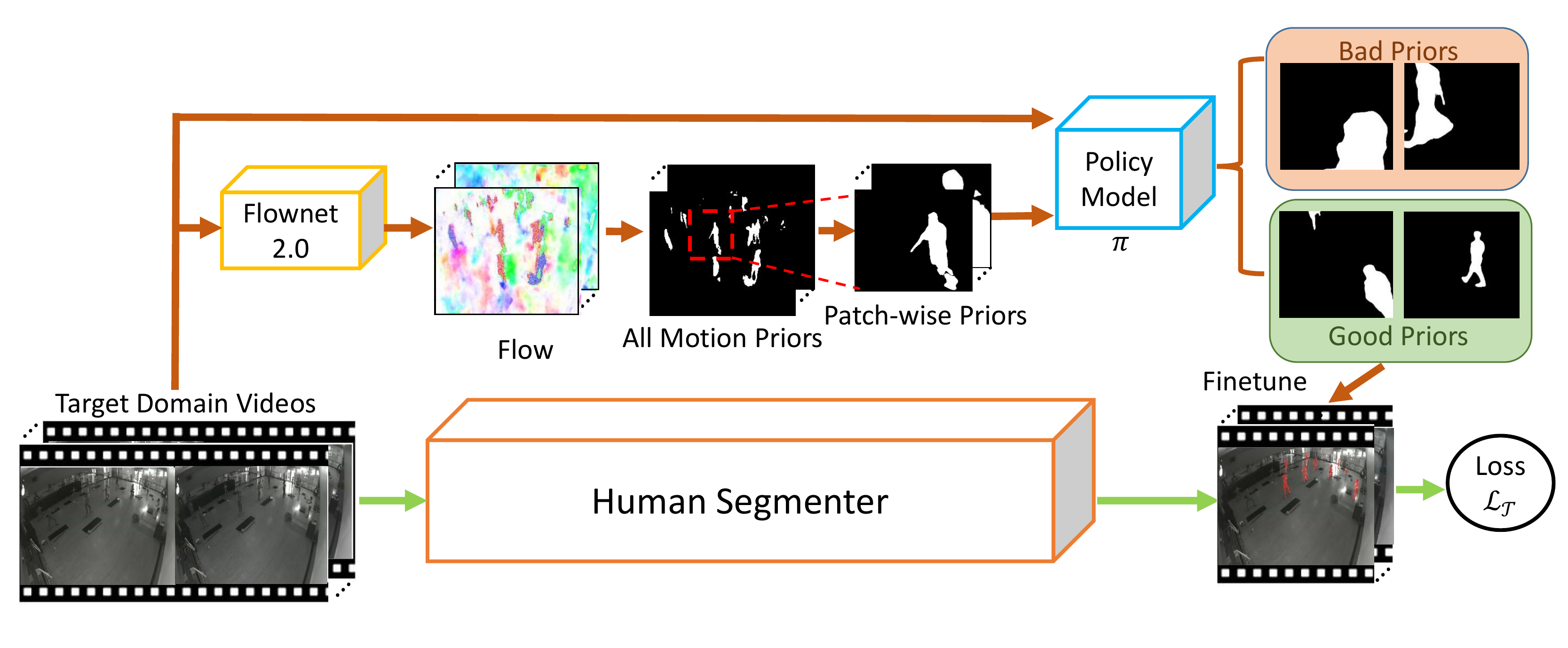}
\end{center}
\caption{The figure illustrates the extraction and usage of motion prior. Top-half shows the path to generate motion priors from videos, followed by policy model-based selection. Bottom-half shows selected priors for fine-tuning segmenter on target domain.}
\label{finetune-tgt}
\end{figure*}

We apply the trained policy $\pi$ to select a set of image patches $\mathcal{K}_\mathcal{T}$ along with strong motion prior from the unlabeled training frames in the target domain $\mathcal{V}^{\mathcal{T}}_T$.
They are referred to as patch-wise \textit{strong} motion prior as below,
\begin{eqnarray}\label{StrongMP}
\mathcal{K}_\mathcal{T}=\{(k,n);a_{k,n}=1\}~.
\end{eqnarray}
Given $\mathcal{K_T}$, we use $(I_{k,n},\textbf{m}(I_k)_n)_{k\in \mathcal{K_T}}$ as additional pairs of image and ground truth human segmentation and introduce cross-entropy loss for fine-tuning in the target domain. See Fig.~\ref{finetune-tgt}.

\section{Experiments} \label{sec:results}


We conduct experiments to validate the proposed weakly-supervised active learning method in cross-modalities and cross-scenes settings. Firstly, the result shows that the proposed policy-based active learning method can select informative samples on a new target domain in Sec.~\ref{exp:pal}. Moreover, we show the proposed active learning method is complementary to recent adversarial-based domain adaptation frameworks~\cite{bousmalis2016domain,chen2017no}. The performance gains of our method integrated with domain adaptation methods are shown in Sec.~\ref{exp:ada}.     

We demonstrate the weakly-supervised active learning with the cross-domains setting via our collected source datasets \textit{Gym} and \textit{Store} in camera modality-RGB, along with multiple target datasets, including our remaining datasets in camera modality-IR, and one public available pedestrian dataset, \textit{UrbanStreet}~\cite{fragkiadaki2012two}, \highlight{which contains 18 stereo sequences of pedestrians taken from a stereo rig mounted on a car driving in the streets of Philadelphia.} 


\subsection{Implementation Details}
In all experiments,  we use U-Net structure~\cite{ronneberger2015u} as our baseline segmentation model for comparison. The code and models are evaluated in the Pytorch framework. For fair comparisons, we use the Intersection over Union (IoU)~\cite{everingham2015pascal} as evaluation metrics for all experiments, where $\text{IoU}=\frac{TP}{TP+TF+FP}$. The quantitative results in Tables.~\ref{result:AL} and~\ref{result:ADA} show the IoU scores of foreground class. For training our policy model, we use initial learning rate of $1\times10^{-4}$ with Adam optimizer~\cite{kingma2014adam}. The discount factor for policy gradient is set to 1. We train about 5000 episodes. In the training procedure, an initialized segmenter pre-trained on MSCOCO~\cite{lin2014microsoft} is further fine-tuned with the policy model. 


\begin{table}[t!]
\caption{
Cross-domain human segmentation performance (IoU) comparison of the proposed weakly-supervised active learning method ``PAL'' with other strategies. \highlight{U- and Seg- denote the model architectures: U-Net and SegNet, respectively.} First row ``Source Only'' is direct application of pre-trained model on target domain data. To best of our knowledge, none of the existing active learning algorithm use only prior instead of true label for fine-tuning on target domain.}
\begin{center}
\scalebox{0.73}{
\begin{tabular}{l | c  c c c c c}
\hline
Source & Gym-RGB  &Gym-RGB & Gym-RGB & Store-RGB & Store-RGB & Store-RGB\\
Target & Gym-IR & Multi-Scene-IR & UrbanStreet(-RGB) & Gym-IR & UrbanStreet(-RGB) & Multi-Scene-IR\\
\hline
\textbf{Source Only }(U-) & 48.6\% & 16.8\% & 48.5\% &26.7\% &61.7\%& 29.2\%\\
\textbf{~ ~ ~ ~ ~ ~ ~ ~ ~}(Seg-) & 51.1\%& 23.6\% & 52.3\% & 23.6\% &63.5\% & 35.8\%\\
\hline
\textbf{PAL ~ ~ ~ ~ ~ }(U-) &55.6\% & 30.5\% & 51.2\% & \textbf{32.3\%} & 64.8\% & 34.3\%\\
\textbf{~ ~ ~ ~ ~ ~ ~ ~ ~}(Seg-) &\textbf{57.0\%}& \textbf{38.4\%} & \textbf{56.6\%} &26.9\% &\textbf{65.3\%}&\textbf{39.0\%}\\
\hline
\textbf{Random ~ ~ ~}(U-) &52.5\% & 26.5\%	& 49.3\% & 29.3\% & 62.4\% & 30.2\%\\
\textbf{~ ~ ~ ~ ~ ~ ~ ~ ~}(Seg-) &56.7\% & 37.2\% & 55.3\% & 24.8\% &63.4\% &33.2\%\\
\hline\hline
\textbf{Human- ~ } ~~~(U-) &57.5\% & 34.6\%&55.8\% & 32.5\% &68.5\% & 41.0\% \\
\textbf{Selection} ~~~~~(Seg-) &57.5\% & 42.3\% & 59.7\% & 32.7\%&65.9\%&46.5\% \\
\hline
\end{tabular}}
\end{center}
\label{result:AL}
\end{table}

\subsection{Weakly-supervised Active Learning with Cross-Domain Setting}\label{exp:pal}
We compare our Policy-based Active Learning method (referred to as \textit{PAL}) with two methods: \emph{Random} and \emph{Human Selection} in Table.~\ref{result:AL}. The number of used motion-prior patches is pre-defined in all settings as a budget $b = 60$. Note that all methods share the same motion prior candidates (cropped patches).

\noindent\textbf{Random.} Randomly select a set of motion priors from a data pool. And we report the average results over ten selected sets.

\noindent\textbf{Human Selection.} \sunmin{We manually select a set of motion priors whose motion priors are closer to true annotations while also considering data divergence. The results can be viewed as an upper bound for our method.}

\begin{figure*}[!t]  
 \begin{center}
\includegraphics[width=12cm]{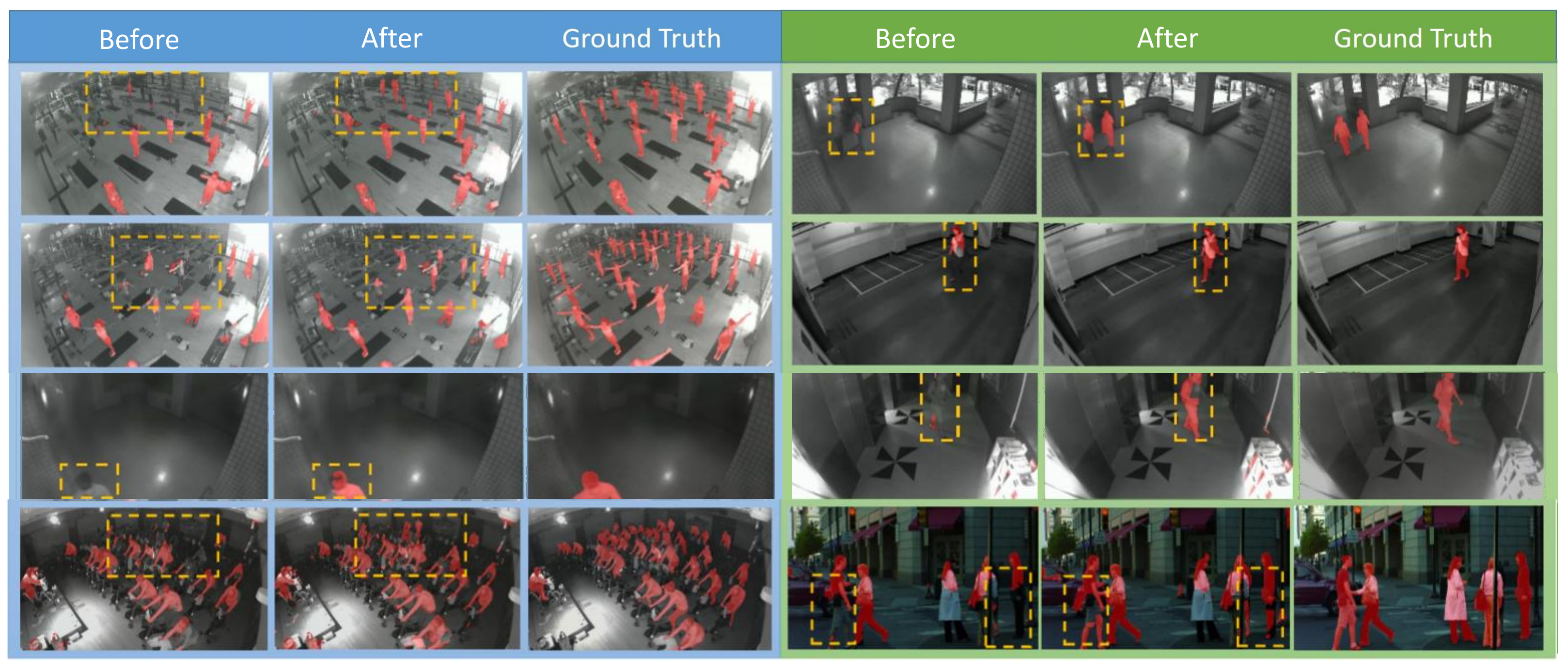}
 \end{center}
\caption{Qualitative results of improving human segmentation on target domain of the following five source-target settings: Store-RGB$\rightarrow$Gym-IR (top-left 6 images), Gym-RGB$\rightarrow$Multi-Scene-IR (top-right 6 images), and Store-RGB$\rightarrow$Multi-Scene-IR (the third row). The last row shows Gym-RGB$\rightarrow$Gym-IR and Gym-RGB$\rightarrow$UrbanStreet, respectively. The columns ``After'' denotes improved segmentations by \textbf{PAL+NMD}. Bounding-boxes in dash-line highlight the significant change.}
 \label{typical}
\end{figure*}

We conduct three kinds of cross-domains applications: (1) cross-modalities, (2) cross-scenes, and (3) cross-modalities \& -scenes. The experimental results are summarized in Table. \ref{result:AL}. \highlight{We choose two baseline segmentation models, U-Net and SegNet, to demonstrate generalization of the method.} We also provide qualitative results in Fig.~\ref{typical}.  

\noindent\textbf{Cross-modalities in same scene.} In our experiment, we change data in Gym from RGB images to infrared images. In Table. \ref{result:AL}, the first column (Gym-RGB to Gym-IR) shows our method ``PAL'' has $+3.1\%$ IoU performance related to random selection and 
improves $+7\%$ IoU from ``Source Only'' (not using information on target domain). 

\noindent\textbf{Cross-scenes in same modality.} 
We also validate our proposed method on public available datasets. However, it's hard to find a public dataset with IR videos with segmentation annotations. We replace with a public dataset \textit{UrbanStreet} as the target domain whose appearance is very different from our surveillance camera dataset but captured in same modality (RGB).
Our method still works under the condition of great appearance change. We conduct two experiments: Gym-RGB $\rightarrow$ UrbanStreet and Store-RGB $\rightarrow$ UrbanStreet showed in Table. \ref{result:AL}. The results show $+2.7\%$ and $+3.1\%$ relative IoU form source model, respectively. 
\sunmin{Note that UrbanStreet contains many moving vehicles. Our method still can distinguish human motion segments form another moving segments, which may come from cars or slight camera motions. This result demonstrates the robustness of our weakly-supervised active learning approach.}

\noindent\textbf{Cross-scenes and -modalities.} This is the most general situation to deal with for applications of surveillance cameras. We show the results of Gym $\rightarrow$ Multi-scene, Store $\rightarrow$ Gym and Store $\rightarrow$ Multi-Scene in Table \ref{result:AL}. Note that all settings are from RGB to IR. 
\sunmin{In all settings, the result shows that PAL offers significant improvement from ``Source Only'' and better than ``Random''. In the case of Store-RGB $\rightarrow$ Gym-IR, the result of our method is very close to the upper bound ``Human Selection'' with only a $0.2\%$ gap.
}
\begin{figure}[t!]
\begin{minipage}{0.55\textwidth} 
\scalebox{0.96}{
\includegraphics[width=6.8cm,height=3cm]
{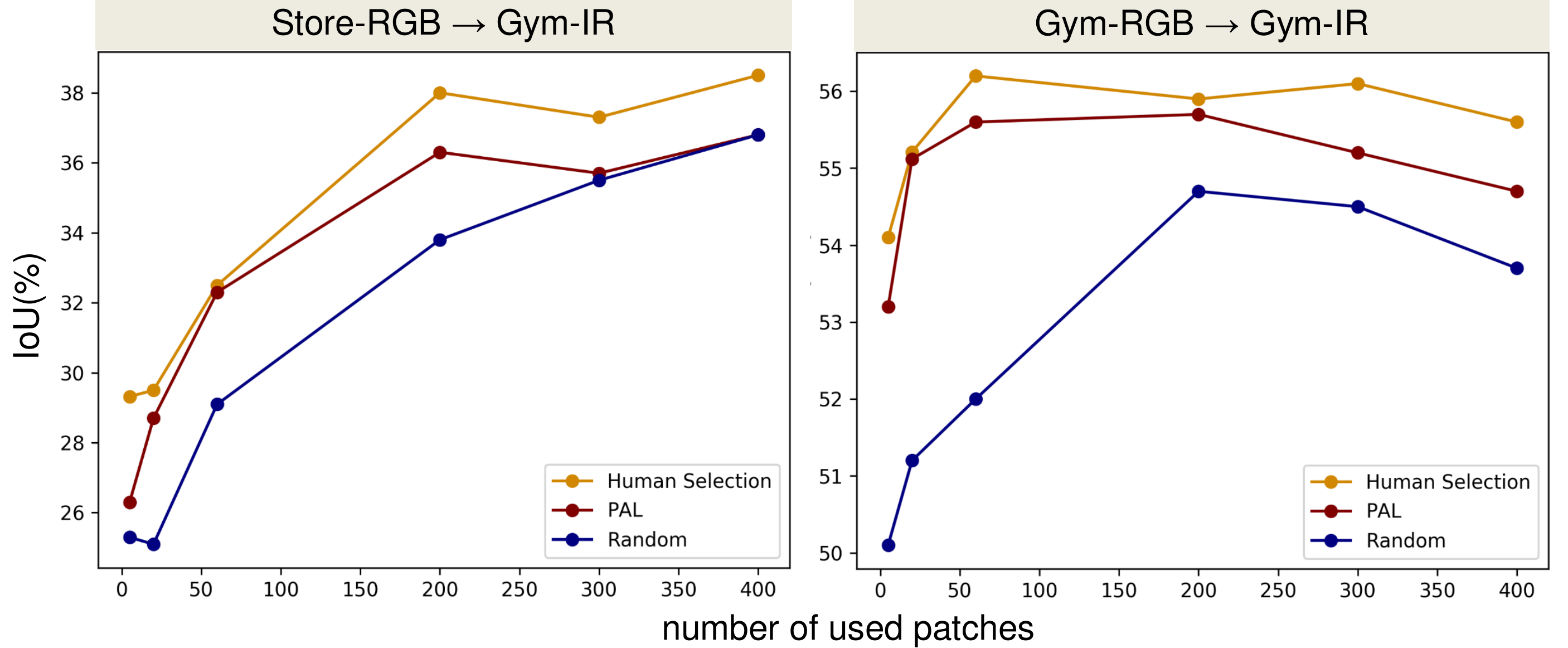}
}
\end{minipage}
\begin{minipage}{0.45\textwidth} 
\caption{The performance of human segmentations on target domain using our \textit{PAL} method, where the policy-based active learning is trained on Gym-RGB and Store-RGB (Source), respectively, and is applied to Gym-IR (Target). Note that only motion prior (ZERO label) is used for  target domain.} 
\end{minipage}
\label{fig:curve}
\end{figure}

\sunmin{The performance curves by exploring incrementally more amounts of priors are shown in Fig.~\ref{fig:curve}. We show the effectiveness of PAL comparing with Random and Human selection results. Interestingly, the curve in Store-RGB $\rightarrow$ Gym-IR implies that the mIoU can increase by adding more strong priors. Since we can obtain motion priors from unlabeled videos with ZERO label cost, our method can be efficient practical to improve performance by simply collecting more unlabeled videos.}


\subsection{Combined with adversarial Domain Adaptation} 
\label{exp:ada}
\begin{table*}[!t]
\caption{Cross-domain human segmentation performance (IoU) comparison of the proposed method (\textbf{bold}) with other baselines in 6 diverse source-target domain pairs. 
The last two rows show the combined methods outperform each of sub-method, implying the active learning approach is complementary to original domain adaptation framework.} 
\begin{center}
\scalebox{0.76}{
\begin{tabular}{l | c c c c c c}
\hline
Source & Gym-RGB  &Gym-RGB & Gym-RGB & Store-RGB & Store-RGB & Store-RGB\\
Target & Gym-IR & Multi-Scene-IR & UrbanStreet(-RGB) & Gym-IR & UrbanStreet(-RGB) & Multi-Scene-IR\\
\hline
Source Only & 48.6\% & 16.8\% & 48.5\% &26.7\% &61.7\%& 29.2\%\\
\textbf{PAL} &55.6\% & 30.5\%	& 51.2\% & 32.3\% & 64.8\% & 34.3\%\\
DSN ~\cite{bousmalis2016domain}&54.3\% & 25.9\% & 52.6\% & 31.8\% & 62.3\% & 34.4\%\\
NMD~\cite{chen2017no} &52.1\%&26.1\%&	52.1\% &31.7\%	&63.1\%	&34.5\%\\
\hline
\textbf{PAL+DSN} &\textbf{55.8}\% & 35.8\%	& \textbf{54.5}\% & \textbf{36.4}\% & \textbf{66.2}\% & \textbf{39.0}\%\\
\textbf{PAL+NMD}&55.6\%&\textbf{36.7}\% &\textbf{54.5}\% &34.0\%& 64.6\%	&36.3\%\\
\hline
\end{tabular}
}
\end{center}
\label{result:ADA}
\end{table*}

\sunmin{In this part, we integrate the proposed weakly-supervised active learning with other existing unsupervised domain adaptation (DA) methods for two reasons. Firstly, unsupervised DA shares the same goal of ZERO label cost on target domain. Secondly, intuitively our method should be complementary to unsupervised DA. Most of the unsupervised DA methods only have fine-tuning loss on source domain, since the label is not available on target domain. 
However, our weakly-supervised active learning policy enables fine-tuning on target domain using the policy-selected strong motion priors.}

On the concern of performance and complexity, we combine proposed PAL with two of existing methods, DSN~\cite{bousmalis2016domain} and NMD~\cite{chen2017no}. 
Demonstrating in same cross-domains settings as the previous section, we do the comparison between proposed PAL with these unsupervised domain adaptation baselines, and show these two types approaches (PAL vs. UDA) are complementary with each other since the combined method reach the greatest improvement on target domain. See results in Table. \ref{result:ADA}. \sunmin{For instance, in the setting Gym-RGB $\rightarrow$ Multi-Scene IR (second column), the combined method ``PAL+NMD'' achieve about $6.2\%$ IoU improvements from each sub-approach.}   

\section{Conclusion} \label{sec:Con}
We propose to leverage ``motion prior'' in videos to improve human segmentation with cross-domain setting. 
We propose a memory-network-based policy model to select ``strong'' motion prior through reinforcement learning. The selected segments have high precision and are used to fine-tune the model on target domain. Moreover, the active learning strategy is shown to be complementary to adversarial-based domain adaptation methods. In a newly collected surveillance camera datasets, we show that our proposed method significantly improves the performance of human segmentation across multiple scenes and modalities.

\section{Acknowledgment}
We thank Umbo CV, MediaTek, MOST 107-2634-F-007-007 for their support.


\bibliographystyle{splncs}
\bibliography{egbib}
\end{document}